\documentclass[lettersize,journal]{IEEEtran}
\usepackage{amsmath,amsfonts}
\usepackage{algorithm}
\usepackage{algpseudocode}
\usepackage{array}
\usepackage[caption=false,font=normalsize,labelfont=sf,textfont=sf]{subfig}
\usepackage{textcomp}
\usepackage{stfloats}
\usepackage{mathtools}
\usepackage{url}
\usepackage{verbatim}
\usepackage{amssymb}
\usepackage{bm}
\usepackage{graphicx}
\usepackage{multirow}
\usepackage{cite}
\usepackage{makecell}
\usepackage{graphicx}
\usepackage{textcomp}
\usepackage{array} 
\usepackage{longtable}
\usepackage{booktabs}
\usepackage{float}
\usepackage{algorithm}
\usepackage{algpseudocode}
\usepackage{amsmath}
\usepackage{algorithmicx}
\usepackage{orcidlink}
\usepackage{xr}

\begin{document}


\title{History-Augmented Contrastive Learning With Soft Mixture of Experts for Blind Super-Resolution of Planetary Remote Sensing Images}

\author{Hui-Jia Zhao \orcidlink{0009-0001-3369-5051}, \textit{Student Member, IEEE}, Jie Lu,Yunqing Jiang, Xiao-Ping Lu* \orcidlink{0000-0002-2363-4175},  Kaichang Di
\thanks{This work was supported in part by the Science and Technology Development Fund, Macau SAR under Grants 0096/2022/A and 0026/2025/RIA1, and in part by the National Key Research and Development Program of China under Grant 2022YFF0503100. (Corresponding author: Xiaoping Lu.)}

\thanks{Hui-Jia Zhao is with the School of Computer Science and Engineering, Macau University of Science and Technology, Taipa, Macao, China (e-mail:3240001399@student.must.edu.mo).}

\thanks{Jie Lu is with the School of Computer Science and Engineering, Macau University of Science and Technology, Taipa, Macao, China (e-mail:jiel13@student.must.edu.mo).}

\thanks{Yunqing Jiang is with the School of Computer Science and Engineering, Macau University of Science and Technology, Taipa, Macao, China (e-mail:3240007841@student.must.edu.mo).}

\thanks{Xiao-Ping Lu is with the School of Computer Science and Engineering, Macau University of Science and Technology, Taipa, Macao, China, and also with the State Key Laboratory of Lunar and Planetary Sciences, Macau University of Science and Technology, Taipa, Macao, China (e-mail: xplu@must.edu.mo).}

\thanks{Kaichang Di is with the State Key Laboratory of Remote Sensing Science, Aerospace Information Research Institute, Chinese Academy of Sciences, Beijing 100101 China (e-mail: dikc@aircas.ac.cn).}}
\markboth{Journal of \LaTeX\ Class Files,~Vol.~14, No.~8, August~2021}%
{Shell \MakeLowercase{\textit{et al.}}: A Sample Article Using IEEEtran.cls for IEEE Journals}

\IEEEpubid{0000--0000/00\$00.00~\copyright~2021 IEEE}

\maketitle

\begin{abstract}
Blind Super-Resolution (BSR) in planetary remote sensing constitutes a highly ill-posed inverse problem, characterized by unknown degradation patterns and a complete absence of ground-truth supervision. Existing unsupervised approaches often struggle with optimization instability and distribution shifts, relying on greedy strategies or generic priors that fail to preserve distinct morphological semantics. To address these challenges, we propose History-Augmented Contrastive Mixture of Experts (HAC-MoE), a novel unsupervised framework that decouples kernel estimation from image reconstruction without external kernel priors. The framework is founded on three key innovations: (1) A Contrastive Kernel Sampling mechanism that mitigates the distribution bias inherent in random Gaussian sampling, ensuring the generation of plausible kernel priors via similarity constraints; (2) A History-Augmented Contrastive Learning strategy that leverages historical model states as negative self-priors. We provide a theoretical analysis demonstrating that this mechanism induces strong convexity in the feature space, thereby stabilizing the unsupervised optimization trajectory and preventing overfitting; and (3) A Morphology-Aware Soft Mixture-of-Experts (MA-MoE) estimator that dynamically modulates spectral-spatial features to adaptively reconstruct diverse planetary topographies. To facilitate rigorous evaluation, we introduce Ceres-50, a benchmark dataset encapsulating diverse geological features under realistic degradation simulations. Extensive experiments demonstrate that HAC-MoE achieves state-of-the-art performance in reconstruction quality and kernel estimation accuracy, offering a solution for scientific observation in data-sparse extraterrestrial environments. The code is available at \url{https://github.com/2333repeat/HAC-MoE}, and the dataset is available at \url{https://github.com/2333repeat/Ceres-50}.
\end{abstract}


\begin{IEEEkeywords}
Meta-learning, contrastive learning, blind super-resolution, planetary remote sensing
\end{IEEEkeywords}



\section{Introduction}
\label{Intro}

\IEEEPARstart{P}{lanetary} remote sensing constitutes the primary observational interface for extraterrestrial exploration, serving as the fundamental basis for  geological morphology analysis, and the investigation of solar system evolution. Unlike terrestrial imaging, where acquisition conditions are often controllable and ground truth is obtainable, planetary imaging operates under an extreme regime of blind inverse problems. Spacecraft are subject to severe payload restrictions, limiting optical aperture size, while data transmission over astronomical distances necessitates aggressive compression. Furthermore, the imaging environments are hostile and highly variable. The strong radiation belts of Jupiter and Saturn induce complex instrumental noise that complicates remote sensing of their icy moons. Consequently, the received images are often low-resolution (LR) and corrupted by unknown, heterogeneous degradation processes, severely impeding downstream scientific analysis.

To recover high-frequency details from these degraded observations, Super-Resolution (SR) techniques have been extensively deployed. While deep learning has revolutionized SR in terrestrial domains \cite{chen2024frequency,chen2025frequency,SRCNN,VDSR,EDSR,SRNO,LIIF}, the direct transfer of these paradigms to planetary exploration remains fundamentally obstructed by the lack of supervision. Standard SR methods rely on paired LR-HR datasets trained on known degradations (typically bicubic downsampling) \cite{zhao2024neural}. However, real planetary degradation is strictly blind, the blur kernels are unknown, spatially variant, and physically complex. We mathematically formulate this Blind Super-Resolution (BSR) problem as:

\begin{equation}
\mathbf{y} = (\mathbf{x} \otimes \mathbf{k}) \downarrow _{s} + \mathbf{z} ,  
\end{equation}

\IEEEpubidadjcol  

\noindent where $\mathbf{y}\in\mathbb{R}^{H\times W}$ denotes the observed LR measurement, $\mathbf{x}\in\mathbb{R}^{sH\times sW}$ is the latent High-Resolution (HR) signal, $\otimes$ represents the convolution operation with an unknown blur kernel $\mathbf{k}\in\mathbb{R}^{K\times K}$, $\downarrow_{s}$ denotes downsampling by scale factor $s$, and $\mathbf{z}$ represents additive noise. The challenge of solving Eq. (1) in a planetary remote sensing images is  highlighting the following limitations:

\textbf{1) The Absence of Ground Truth Supervision:} Most existing BSR frameworks, even those estimating kernels, ultimately rely on large-scale paired datasets or pretrained priors derived from natural images (e.g., DIV2K) \cite{liu2022Blind_image_super_resolution_A_survey_and_beyond,RCAN,USRnet,SRMD,DnCNN}. In planetary science, ground truth HR images simply do not exist. This necessitates a purely unsupervised approach, yet removing supervision typically leads to optimization instability.

\begin{figure}[t]  
\centering  
\includegraphics[width=3.5in]{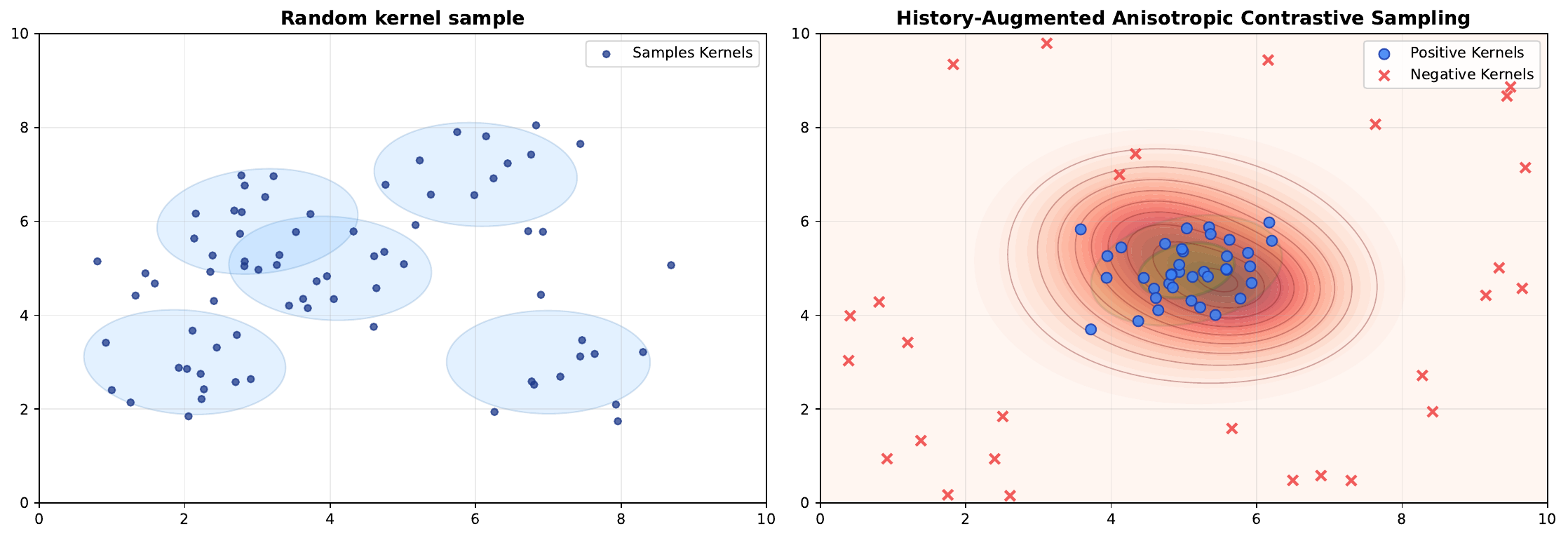}  
\caption{ Comparison of kernel distribution between random kernel samples and contrastive kernel sampling. The contrastive kernel sampling effectively generates kernels with balanced distribution. It also reduces the probability of generating outlier kernels, which can adversely affect image reconstruction.} 
\label{kernel_distribution_comparison}  
\end{figure}

\textbf{2) Distributional Mismatch in Kernel Estimation:} Existing unsupervised methods often employ alternating optimization strategies that require sampling degradation kernels to train the image reconstructor \cite{MLMC,DDSR,liang21FKP_DIP}. However, these methods typically sample kernel parameters from uniform or simple Gaussian distributions. As illustrated in Figure \ref{kernel_distribution_comparison}, there is a significant distribution shift in random sampled kernels. Random sampling often generates outlier kernels that  causing the image reconstruction network to overfit to artifacts rather than learning valid kernel priors.

\textbf{3) The planetary remote sensing domain lacks BSR datasets:} The lack of planetary remote sensing datasets significantly hinders development in this area. The evaluation of planetary remote sensing image restoration tasks also requires BSR datasets.

To address these fundamental disconnects, we propose History-Augmented Contrastive Mixture of Experts (HAC-MoE), a novel unsupervised BSR framework tailored for the data-sparse, degradation-heavy domain of planetary observation. HAC-MoE decouples the ill-posed problem into two robustly regularized sub-problems: kernel estimation and image reconstruction. Specifically, we introduce a Contrastive Kernel Sampling mechanism that replaces naive random sampling. By utilizing contrastive learning, we force the generated kernels to cover a diverse but plausible, rejecting outlier kernels that disrupt training stability. Furthermore, we propose a History-Augmented Contrastive Learning strategy. The current model states are consistent with their historical moving averages. By treating historical model states as negative self-priors, we induce a regularization term that effectively prevents the overfitting common in unsupervised DIP-based methods. Finally, recognizing that planetary surfaces exhibit vast morphological diversity (e.g., smooth plains vs. high-frequency craters), we design the image estimator as a Morphology-Aware Soft Mixture-of-Experts (MA-MoE). This architecture dynamically routes features based on local spectral-spatial complexity, ensuring that the reconstruction logic adapts to specific geological textures.

The primary contributions of this work are summarized as follows:
\begin{enumerate}
    \item We propose HAC-MoE, the first unsupervised BSR framework specifically designed for planetary remote sensing, capable of operating without any paired training data or external kernel priors.
    \item We propose a History-Augmented Contrastive mechanism that enforces temporal consistency in the optimization trajectory. We provide a theoretical analysis demonstrating that this mechanism improves convergence stability in the feature space compared to standard unsupervised baselines.
    \item We propose the Contrastive Kernel Sampling to resolve the unbalanced distribution caused by random kernel sampling. It can reduce the probability of generating outlier kernels.
    \item We propose Ceres-50, a benchmark dataset comprising high-quality observations of the dwarf planet Ceres, facilitating standardized evaluation for the planetary task.
\end{enumerate}

\section{Related Work}

\subsection{Non-blind Super-resolution}

Non-blind super-resolution focuses on reconstructing images under bicubic degradation \cite{tcsvt_sr_1,tcsvt_sr_2}. SRCNN was the first deep learning-based end-to-end super-resolution method \cite{SRCNN}. Residual connections have enabled deep neural networks in the super-resolution domain \cite{EDSR,RDN,RCAN,MSRN,tcsvt_sr_3}. In remote sensing, end-to-end supervised learning and specialized neural network designs have become popular \cite{FeNet_light_weight_remote_sensing_sr,LAM_for_remotesensing}. Additionally, ZSSR achieves unsupervised super-resolution through deep internal learning \cite{zssr}, while meta-learning improves convergence by finding initialization-sensitive weights \cite{mzsr}.

\subsection{Blind Super-resolution}
Unlike non-blind SR, BSR reconstructs images under degradation kernels and additive noise instead of bicubic downsampling, achieving more complex and realistic degradation \cite{liu2022Blind_image_super_resolution_A_survey_and_beyond,tcsvt_bsr_1,tcsvt_bsr_2}. In BSR, kernel estimation plays an important role prior to image reconstruction, with image reconstruction typically conducted after kernel estimation \cite{liu2022Blind_image_super_resolution_A_survey_and_beyond,zhang2021plug}. KernelGAN estimates the degradation kernel through image internal recurrence \cite{kernel_gan}. Decoupling features through contrastive learning to focus on resolution-invariant features while ignoring resolution-variant features also yields strong performance \cite{zhang2021blindSR_Contrastive_Representation_Learning}. Pre-trained BSR models have achieved strong performance in remote sensing applications \cite{BSR_remote_sensing_Hyperspectral,BSR_remote_sensing_unsupervised,multilayer_degradation_BSR,blind_hyperspectral_image,Deep_blind_super_resolution_for_satellite_video}. For unsupervised BSR, Flow-based Kernel Prior (FKP) leverages normalizing flows to learn the mapping between Gaussian kernel distributions and latent variables \cite{liang21FKP_DIP}. Combining meta-learning with random kernel generators also achieves strong performance \cite{DDSR}. Simulating degradation kernels with Markov Chain Monte Carlo and using Langevin dynamics to optimize the network also mitigate the overfitting \cite{MLMC}.

\begin{figure*}[t]
\centering
\includegraphics[width=6.5in]{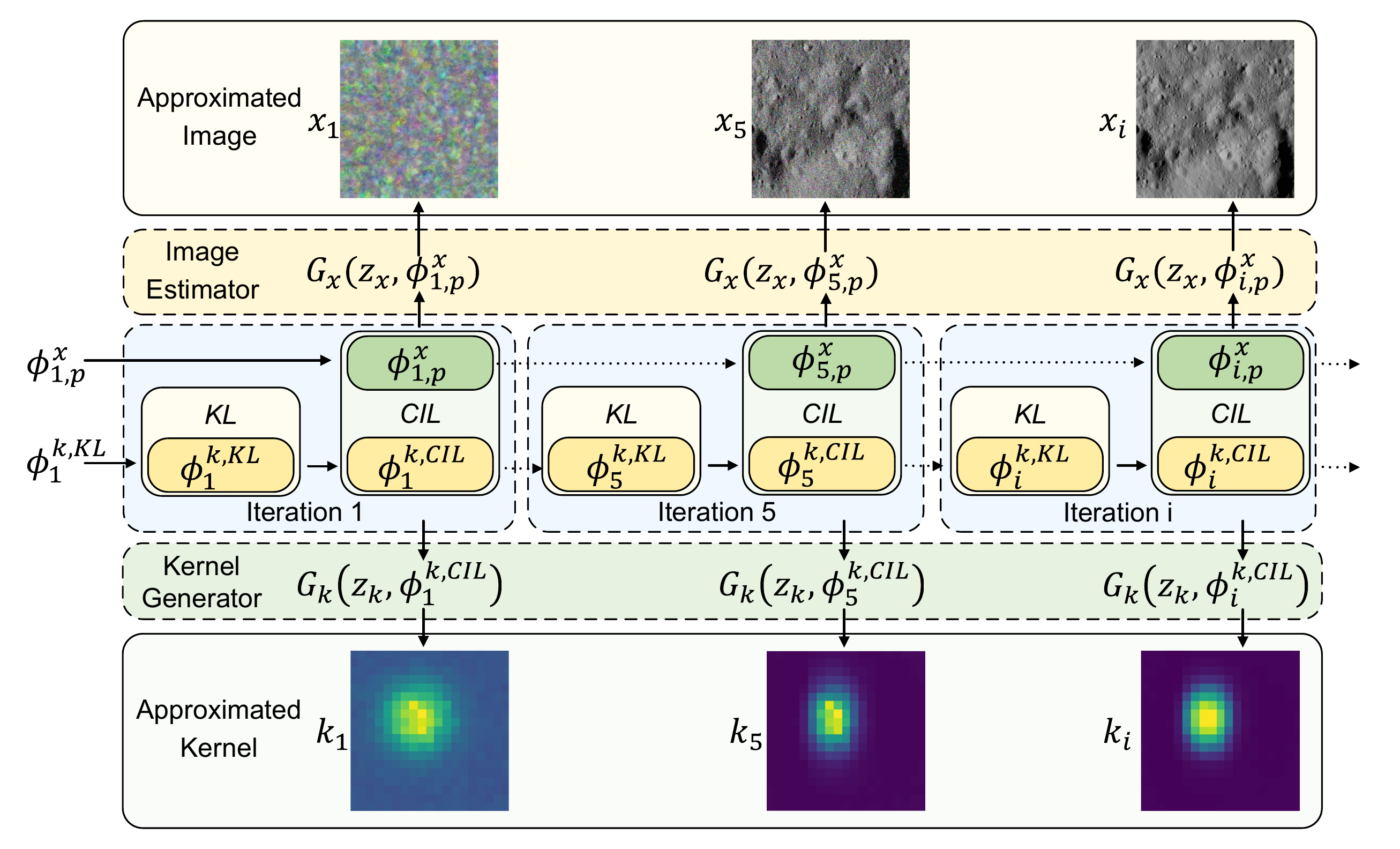}
\caption{Overall workflow of the proposed framework. The Kernel learning (KL) stage and Contrastive Image Learning (CIL) stage will be optimized alternately. In KL stage, the parameter of kernel generator $\phi_{i}^{k,KL}$ will be optimized. In CIL stage, the parameter of kernel generator and image estimator  $\phi_{i}^{k,CIL}$ and $\phi_{i,1}^{x}$ will be optimized. After the $N$-th iteration, the image estimator and the kernel generator generate the SR image $x_{N}$ and the output kernel $k_{N}$ as the final output. The detailed algorithm is presented in Algorithm~\ref{HAC-MoE}.}
\label{workflow}
\end{figure*}

\subsection{Contrastive Learning}
Beyond supervised learning and self-supervised learning, the contrastive learning has achieved strong performance across various tasks \cite{A_comprehensive_survey_on_contrastive_learning}. Contrastive learning focuses on learning common features between similar objects and distinguishing features from dissimilar objects \cite{A_comprehensive_survey_on_contrastive_learning,jaiswal2020survey_ontrastive_self-supervised_learning}. Classifying images based on non-parametric instance similarity achieves strong performance \cite{non_parametric_instance_discrimination}. By constructing a dynamic queue, Momentum Contrast (MoCo) achieves strong performance in unsupervised visual representation learning \cite{moco}. In BSR, generating negative samples from a neural network with historical parameters provides a model-level contrastive learning solution \cite{wu2024learning_from_history}. Extracting resolution-invariant information through contrastive decoupling encoding and achieving feature refinement with contrastive learning also yields strong image reconstruction performance \cite{zhang2021blindSR_Contrastive_Representation_Learning}.

\subsection{Mixture of Experts}


Mixture-of-Experts (MoE) architectures decompose complex mappings into sets of specialized expert networks and gating (or routing) modules that combine their outputs conditioned on the input. Recent surveys provide systematic overviews of modern MoE models, summarizing key components such as gating functions (e.g. soft or sparse top-$k$ routing), expert parametrization, routing and load-balancing mechanisms, training strategies, and system-level optimizations. These surveys demonstrate that MoE offers an effective approach to scaling model capacity under limited computational budgets \cite{mu2025comprehensive,gan2025mixture,dimitri2025survey}. These works also emphasize the importance of expert specialization and diversity, the trade-off between routing sparsity and stability, and the challenges of efficiently deploying MoE at scale \cite{mu2025comprehensive,gan2025mixture,cai2025survey,riquelme2021scaling}. However, existing MoE studies primarily focus on classification or sequence modeling tasks and on token-wise or patch-wise sparse routing \cite{mu2025comprehensive}. The existing MoE work on super-resolution tasks also primarily focuses on sparse routing under supervised learning \cite{emad2022moesr,chen2025heterogeneous,mu2025comprehensive}.


\section{Method}
\label{Method}

\begin{figure*}[t]
\centering
\includegraphics[width=6.5in]{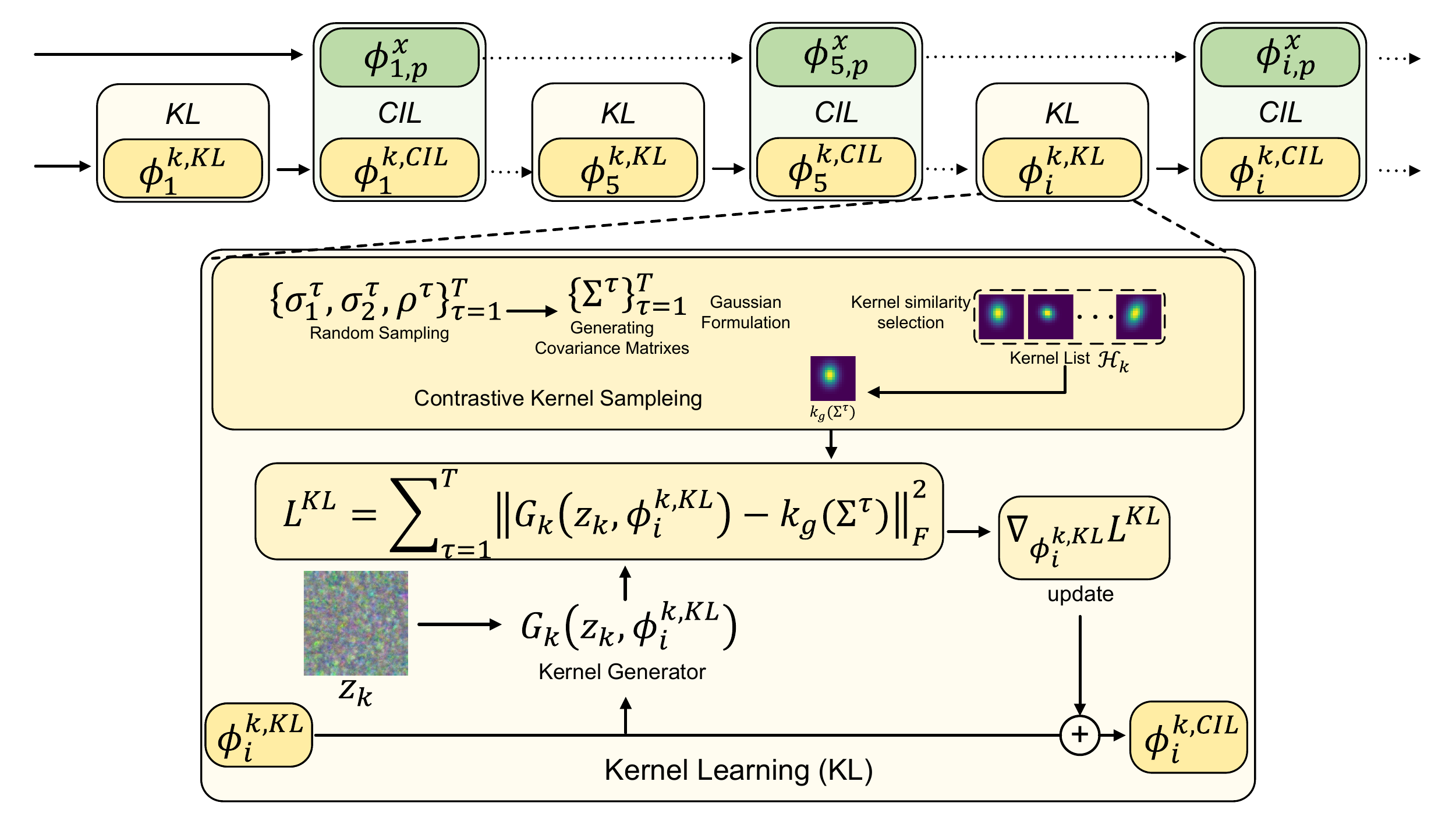}
\caption{Workflow of the Kernel Learning (KL) stage. Contrastive Kernel Sampling generates a queue of diverse, similarity-controlled kernels. The kernel generator $G_k(\cdot)$ is then optimized to map random noise $z_k$ to this  kernel. }
\label{fig:KL_stage}
\end{figure*}

\subsection{Problem Formulation and Framework Overview}
The BSR of planetary images is formally cast as a blind deconvolution and upsampling problem, where the goal is to recover the latent high-resolution (HR) image $\mathbf{x}$ and the degradation kernel $\mathbf{k}$ from a low-resolution (LR) observation $\mathbf{y}$. To address the severe ill-posedness of this unsupervised task, we propose the History-Augmented Contrastive Mixture of Experts (HAC-MoE) framework. 

As illustrated in Figure~\ref{workflow}, HAC-MoE decouples the optimization trajectory into two alternating, mutually reinforcing stages: 
\begin{enumerate}
    \item \textbf{Kernel Learning (KL) Stage:} We introduce a contrastive Kernel Sampling strategy to construct a degradation manifold that is physically plausible yet diverse, preventing the kernel generator $G_k(\cdot)$ from collapsing into trivial or impulsive solutions.
    \item \textbf{Contrastive Image Learning (CIL) Stage:} We employ a Morphology-Aware Mixture-of-Experts (MA-MoE) as the image estimator $G_x(\cdot)$. To stabilize the unsupervised optimization, we incorporate a history-augmented contrastive mechanism that leverages historical model states as negative self-priors, effectively regularizing the solution space.
\end{enumerate}

Formally, let $\phi^{k}$ and $\phi^{x}$ denote the parameters of the kernel generator and image estimator, respectively. The optimization alternates between updating $\phi^{k}$ to minimize the distributional divergence of estimated kernels, and updating $\phi^{x}$ to maximize reconstruction fidelity under the estimated kernel prior.


\subsection{Morphology-Aware Mixture of Experts (MA-MoE)}
\label{sec:MA_MoE}

\begin{figure*}[t]
\centering
\includegraphics[width=6.5in]{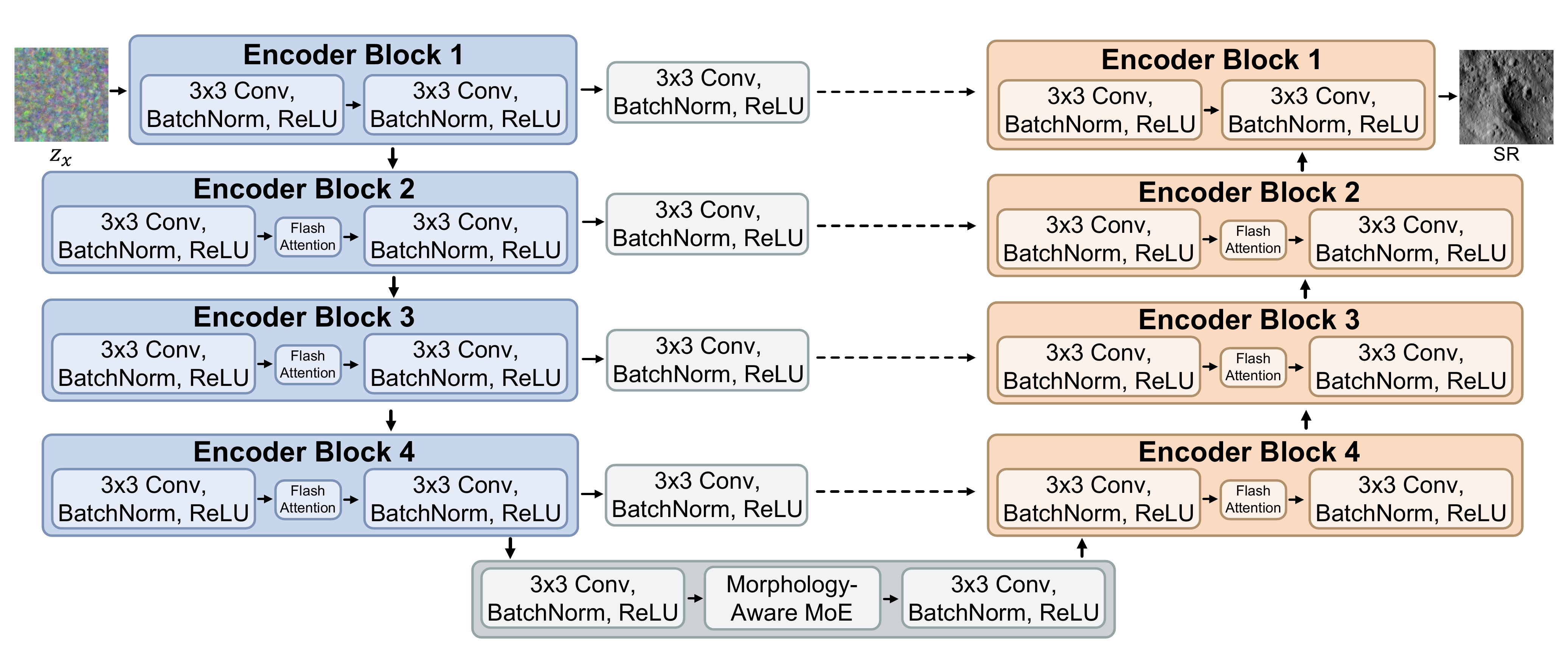}
\caption{The structure of image estimator $G_x(\cdot)$.  The image estimator takes random noise $z_x$ as input and progressively reconstructs the high-resolution image $\hat{x}$ through an encoder-decoder pathway with skip connections.}
\label{unet}
\end{figure*}

\begin{figure}[t]  
\centering  
\includegraphics[width=3.3in]{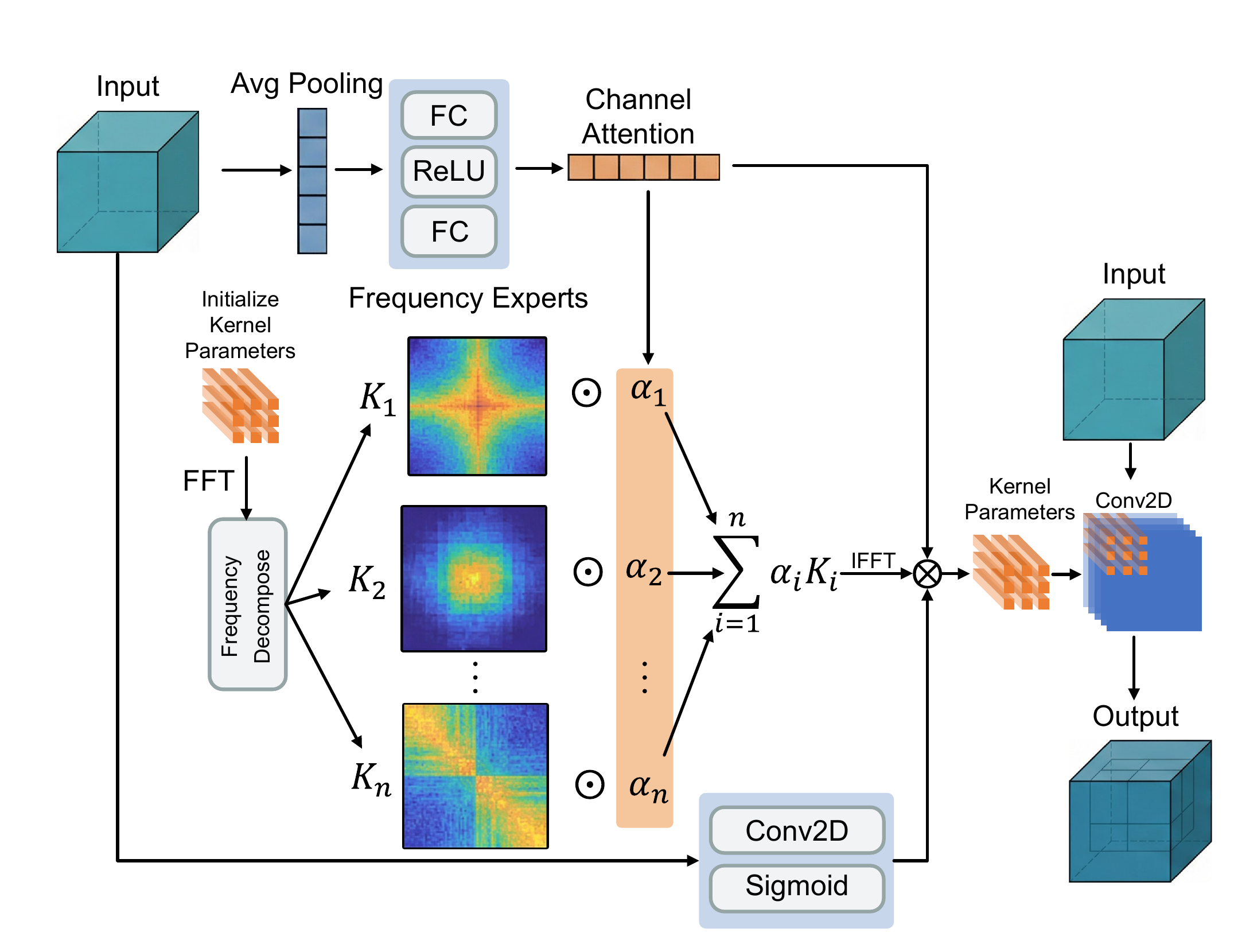} 
\caption{Architecture of the Morphology-Aware Mixture of Experts (MA-MoE). The network dynamically routes spectral features to specific frequency experts based on the local terrain morphology (e.g., craters vs. plains).}  
\label{fig:moe_arch}  
\end{figure}

Planetary topographies exhibit significant morphological heterogeneity: impact craters and tectonic fractures are characterized by high-frequency discontinuities, while volcanic plains and icy regoliths differ primarily in low-frequency spectral signatures. A static convolutional kernel shares weights across the entire spatial domain, inevitably forcing a compromise between sharpening edges and smoothing noise. Following \cite{chen2025frequency}, we propose the MA-MoE (Figure~\ref{fig:moe_arch} and Figure~\ref{unet}), which dynamically modulates feature processing based on local spectral-spatial complexity.

Let $\mathbf{X} \in \mathbb{R}^{C \times H \times W}$ represent the input feature tensor. We define a bank of $K$ spectral experts $\mathcal{E} = \{ \mathbf{E}_1, \mathbf{E}_2, \dots, \mathbf{E}_K \}$, where each expert is implemented as a learnable filter bank specialized for distinct frequency bands. The expert selection is governed by a Morphology Gating Network.

\subsubsection{Morphology Gating Network}
The gating network functions as a global context router. It aggregates global terrain semantics via Global Average Pooling (GAP) and projects them into a selection vector $\alpha \in \mathbb{R}^K$:
\begin{equation}
    \alpha = \sigma \left( \mathcal{F}_{\mathrm{mlp}}(\text{GAP}(\mathbf{X})) \right),
\end{equation}
where $\sigma(\cdot)$ denotes the Sigmoid activation. Unlike sparse Top-$k$ gating, we employ a soft routing mechanism. This allows the network to synthesize a continuous blending of experts, essential for reconstructing complex transition zones (e.g., a crater fading into a smooth plain).

\subsubsection{Dynamic Spectral Aggregation}
The effective convolution kernel $\hat{\mathbf{W}}$ for the current input is synthesized dynamically:
\begin{equation}
    \hat{\mathbf{W}} = \sum_{k=1}^{K} \alpha_k \cdot \mathbf{E}_k.
\end{equation}
To efficiently capture long-range dependencies inherent in geological structures, the modulation is performed in the Fourier domain. The output feature map $\mathbf{Y}$ is computed as:
\begin{equation}
    \mathbf{Y} = \text{IFFT} \left( \hat{\mathbf{W}} \odot \text{FFT}(\mathbf{X}) \right),
\end{equation}
where $\odot$ denotes element-wise multiplication. This formulation allows the MA-MoE to act as an adaptive spectral filter, selectively amplifying high-frequency structural details or suppressing noise in smooth regions based on the inferred terrain morphology.

\subsection{Stage I: Kernel Learning via Contrastive Sampling}
\label{KL_stage}

Existing unsupervised BSR methods typically sample kernel parameters from fixed uniform distributions. As shown in Figure~\ref{kernel_distribution_comparison}, this naive sampling creates a distributional mismatch, generating outlier kernels  that destabilize the image estimator. To address this, we introduce Contrastive Kernel Sampling.

\subsubsection{Contrastive Kernel Sampling}
We treat kernel estimation as a representation learning problem. We maintain a kernel history queue $\mathcal{H}_{\mathrm{k}}=\{(\mathbf{k}_t)\}_{t=1}^{T}$ representing the manifold of physically plausible degradations. For a newly sampled candidate kernel $\mathbf{k}_c$ (generated via random anisotropic Gaussian parameters), we compute a composite similarity score $\mathcal{S}(\mathbf{k}_c, \mathbf{k}_t)$ against the history:
\begin{equation}  
\begin{split}  
\mathcal{S}(\mathbf{k}_{\mathrm{c}},\mathbf{k}_t) = &\lambda_{r}\,\rho(\mathbf{k}_{\mathrm{c}},\mathbf{k}_t) + \lambda_{\mathrm{s}}\,\mathrm{SSIM}(\mathbf{k}_{\mathrm{c}},\mathbf{k}_t) \\ 
&+ \lambda_{f}\,\!\mathbf{f_{s}}(\mathbf{k}_{\mathrm{c}},\mathbf{k}_t),
\end{split}  
\end{equation}  
where $\rho$ is the Pearson correlation, and $\mathbf{f_{s}}$ measures statistical moment similarity. And the $\lambda_{r}$ = 0.5, $\mathrm{s}$ =0.2 and $\lambda_{f}$ =0.3. We define an acceptance criterion $\mathcal{J}(\mathbf{k}_c)$ that rewards diversity (low similarity to immediate neighbors) while penalizing outliers (low similarity to the manifold mean):
\begin{equation}  
\mathcal{J}(\mathbf{k}_{\mathrm{c}}) = -\big|\mathcal{S}_{\mathrm{avg}}-\tau\big| - (\mathcal{S}_{\max}-\sigma_{\max}) - (\sigma_{\min}-\mathcal{S}_{\min}),
\end{equation}  
where $\tau, \sigma_{\max}, \sigma_{\min}$ are hyperparameters controlling the manifold density.  This rejection sampling ensures that the kernel generator $G_k$ is trained only on valid samples.

\subsubsection{Kernel Generator Optimization}
The kernel generator $G_k(\cdot)$, parameterized by $\phi^{k,KL}$, learns to map random noise $z_k$ to the sampled kernel distribution $\{k_g(\Sigma_\tau)\}_{\tau=1}^{T}$. The optimization objective is:
\begin{equation}  
L^{KL} =  { \sum_{\tau=1}^{T}}\left\| G_{k}(z_{k},\phi_{i}^{k,KL}) - k_{g}(\Sigma^{\tau}) \right\|_{F}^{2}.
\end{equation}  
This step effectively pre-warm the kernel estimator, providing a stable degradation prior for the subsequent image reconstruction stage.

\subsection{Stage II: History-Augmented Contrastive Image Learning}
\label{CIL_stage}

\begin{figure*}[!t]
\centering
\includegraphics[width=6.5in]{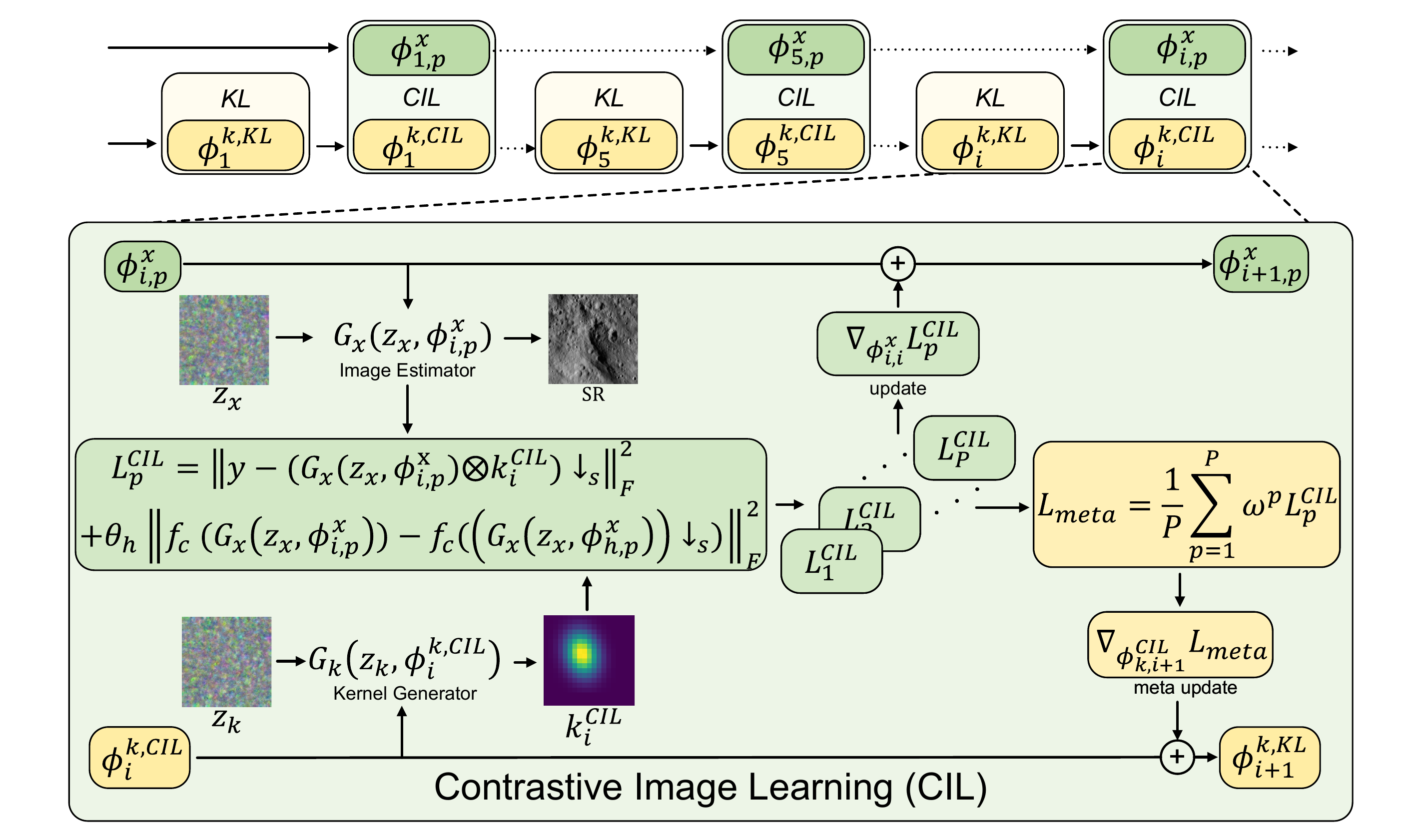}
\caption{Workflow of the Contrastive Image Learning (CIL) stage. The framework optimizes the image estimator $G_x$ and kernel generator $G_k$ under observation consistency. Historical models serve as negative keys in the contrastive loss to prevent overfitting.}
\label{fig:CIL_workflow}
\end{figure*}

In the CIL stage (Figure~\ref{fig:CIL_workflow}), we jointly optimize the image estimator and the kernel generator using the observed LR image $y$. The fundamental challenge here is preventing the network from overfitting to the noise $\mathbf{z}$. We propose a History-Augmented Contrastive Learning mechanism to enforce temporal regularization.

\subsubsection{Observation Consistency and Contrastive Regularization}
Let $\phi_{i,p}^{x}$ denote the parameters of the image estimator at the $p$-th inner-loop iteration of step $i$. We generate a prior using a historical version of the model, parameterized by $\phi_{h,p}^{x}$ (an exponential moving average of past states). The loss function comprises a reconstruction term and a contrastive regularization term:
\begin{equation}  
\label{CIL_formula}
\begin{aligned}
L^{CIL}_{p} &= {\left\| y - (G_{x}(z_{x},\phi_{i,p}^{x})\otimes k^{CIL}_{i}) \downarrow_{s} \right\|^2_{F}} \\
&+ \theta_{h} \cdot {\left\| f_{c}(G_{x}(z_{x},\phi_{i,p}^{x})) - f_{c}(G_{x}(z_{x},\phi_{h,p}^{x})) \right\|^2_{F}},
\end{aligned}
\end{equation}  
where $f_c$ is a feature encoder (CILP \cite{CILP}) extracting semantic features. The contrastive term penalizes the current model if its feature representation diverges significantly from the stable historical trajectory, effectively damping high-frequency oscillations and preventing the over-sharpening of noise.

\begin{figure}[H]  
\centering  
\includegraphics[width=3.3in]{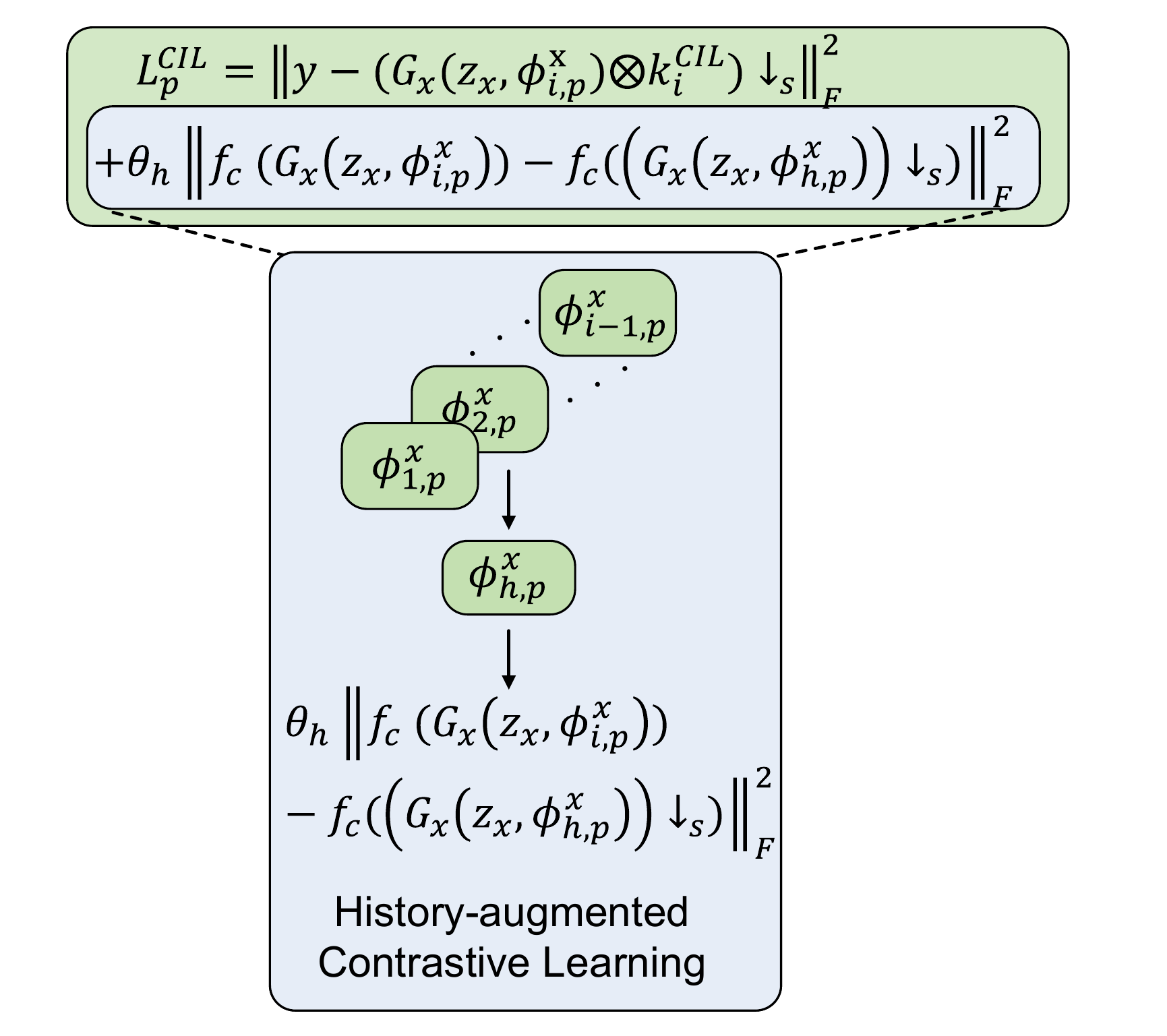} 
\caption{Concept of History-Augmented Optimization. By anchoring the current update to the historical manifold, we induce stronger convexity in the loss landscape, avoiding local minima associated with noise overfitting.}  
\label{fig:history_concept}  
\end{figure} 

\subsubsection{Meta-Learning for Robust Kernels}
To prevent the kernel generator from greedily adapting to a specific image instance (which would degrade generalization), we employ a meta-learning update strategy. We define a meta-loss $L_{Meta}$ as a weighted average of the reconstruction losses over $P$ inner steps:
\begin{equation}  
L_{Meta} = \frac{1}{P} \sum_{p=1}^{P} \omega^p  L^{CIL}_{p}.
\end{equation}  
The weights $\omega^p$ are computed adaptively based on the difficulty of the reconstruction step (using a logarithmic re-weighting of the normalized loss $\pi_p$), ensuring the kernel generator focuses on the most informative gradient directions. The kernel parameters are updated via:
\begin{equation}  
\phi_{i+1}^{k,KL} \leftarrow \phi_{i}^{k,CIL} - \gamma_{k}^{CIL} \cdot \mathrm{Adam}\!\left(\frac{\partial L_{\mathrm{Meta}}}{\partial \phi_{i}^{k,CIL}}\right).
\end{equation} 

\subsubsection{Adaptive History Weighting}
\label{History Contrastive Learning}
The strength of the historical anchor is dynamically adjusted. We employ a loss-adaptive momentum $\alpha$ for updating the historical parameters $\phi_{h}^{x}$:
\begin{equation}  
\alpha = \alpha_{\max} - \frac{L_{\text{curr}} - L_{\min}}{L_{\max} - L_{\min}} (\alpha_{\max} - \alpha_{\min}).
\end{equation}  
When the reconstruction error $L_{\text{curr}}$ is high (early training or difficult terrain), $\alpha$ decreases, allowing the model to learn more rapidly from the current data. As the loss converges, $\alpha$ increases, enforcing consistency. And the expected range $[L_{\min}, L_{\max}] = [10^{-6}, 10^{-2}]$, and the weight range is $[\alpha_{\min}, \alpha_{\max}] = [0.8, 0.99]$. The historical parameters update as:
\begin{equation}  
\phi_{h,p+1}^{x}\leftarrow \alpha \phi_{h,p}^{x} + (1-\alpha) \phi_{i,p}^{x}.
\end{equation}  

The complete optimization procedure is summarized in Algorithm~\ref{HAC-MoE}.

\begin{algorithm}[H]
\caption{History-Augmented Contrastive Mixture of Experts (HAC-MoE)}
\label{HAC-MoE}
\begin{algorithmic}[1]
    \State \textbf{Input}: Low-resolution planetary image $y$. 
    \State \textbf{Initialize}: Noise vectors $z_{k}, z_{x}$; Parameters $\phi_{1}^{k,KL}, \phi_{1,1}^{x}$.
    \For{$i \leftarrow 1$ to $N$}
        \State //  Stage I: Kernel Learning 
        \State Generate manifold samples $\{k_{g}( \Sigma^{\tau} )\}_{\tau=1}^{T}$ via Contrastive Kernel Sampling.
        \State Optimize kernel prior: $\phi_{i}^{k,CIL} \leftarrow \text{Optimize}(L^{KL})$.
        \State Estimate kernel: $k_{i}^{CIL} \leftarrow G_{k}(z_{k},\phi_{i}^{k,CIL})$.
        
        \State // Stage II: Contrastive Image Learning 
        \For{$p \leftarrow 1$ to $P$}
            \State Compute $L^{CIL}_{p}$ via Eq. (\ref{CIL_formula}) using MA-MoE $G_x$.
            \State Update image estimator $\phi_{i,p+1}^{x}$ via backpropagation.
            \State Update history $\phi_{h}^{x}$ via adaptive momentum.
        \EndFor
        \State Compute meta-weights $\omega^p$ and meta-loss $L_{\mathrm{Meta}}$.
        \State Update kernel generator $\phi_{i+1}^{k,KL}$ via $\nabla L_{\mathrm{Meta}}$.
        \State $\phi_{i+1,1}^{x} \leftarrow \phi_{i,P}^{x}$
    \EndFor
    \State \textbf{Output}: Final Kernel $k_{N}$, SR Image $x_{N}$.
\end{algorithmic}
\end{algorithm}

\begin{figure*}[!t]  
\centering  
{\includegraphics[width=6.5in]{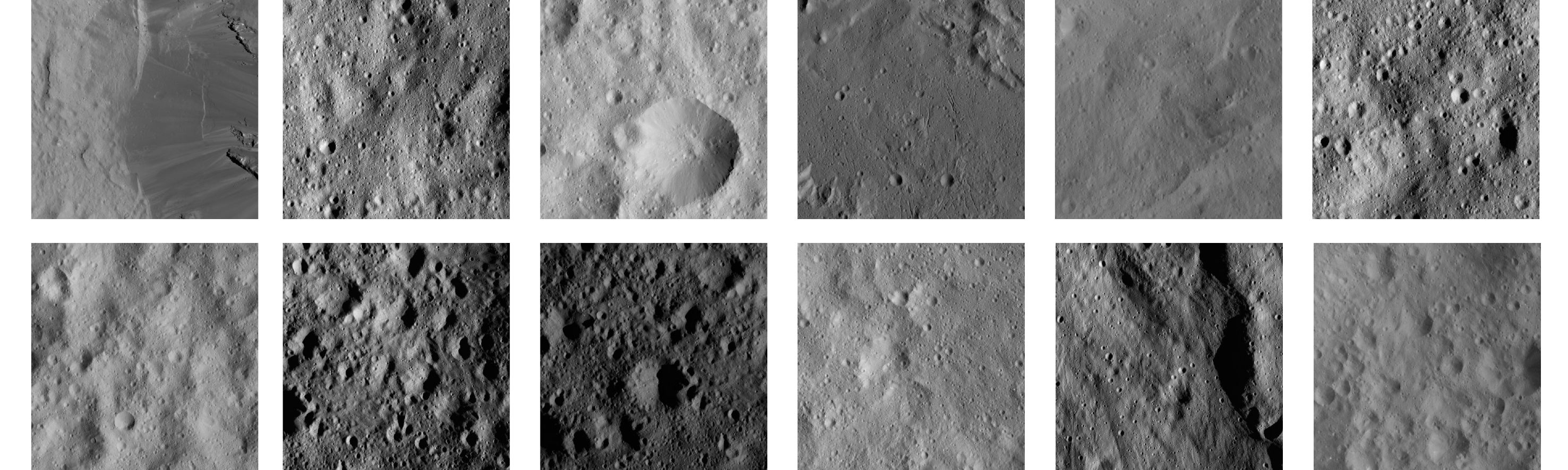}}  
\caption{Example images of the Ceres-50 dataset.}  
\label{dataset}  
\end{figure*}

\begin{figure*}[!t]
\centering
{\includegraphics[width=7in]{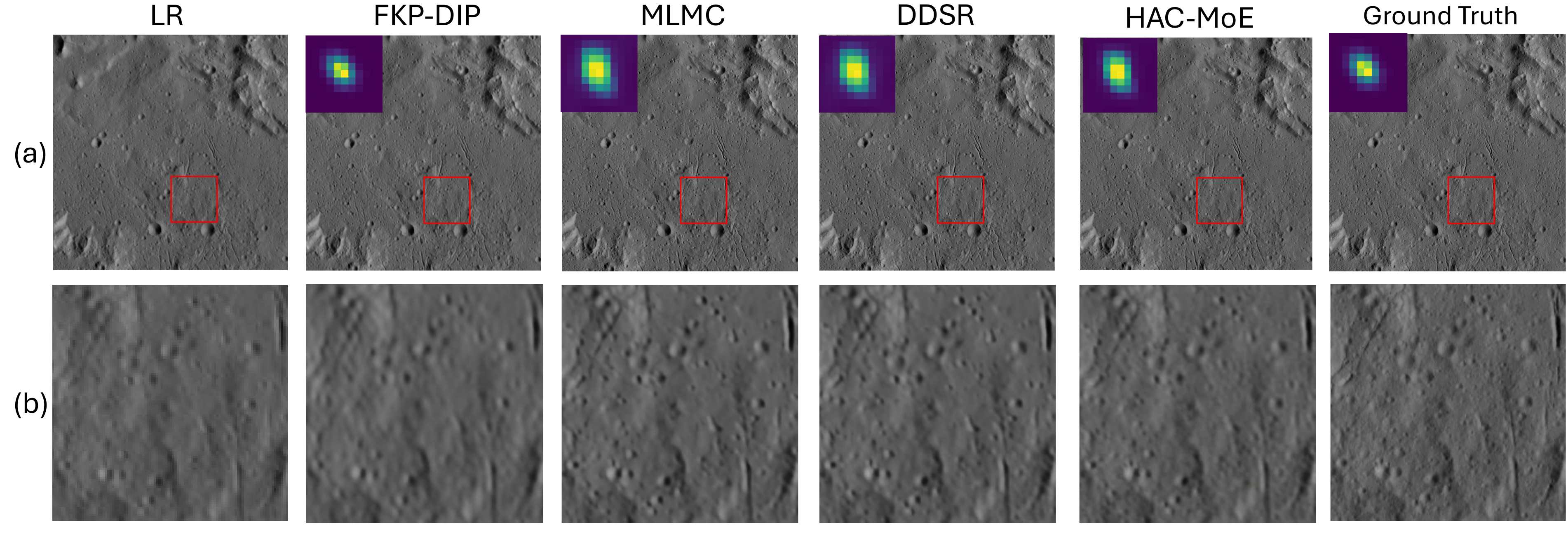}}
\label{fig_first_case}
\caption{Perceptual comparisons of different methods in 2$\times$. The (a) include the ground truth, LR and reconstructed images, the estimated kernels are shown on top left. The (b) denote the zoomed-in for the select area from (a).}
\label{fig2x}
\end{figure*}

\begin{figure*}[!t]
{\includegraphics[width=7in]{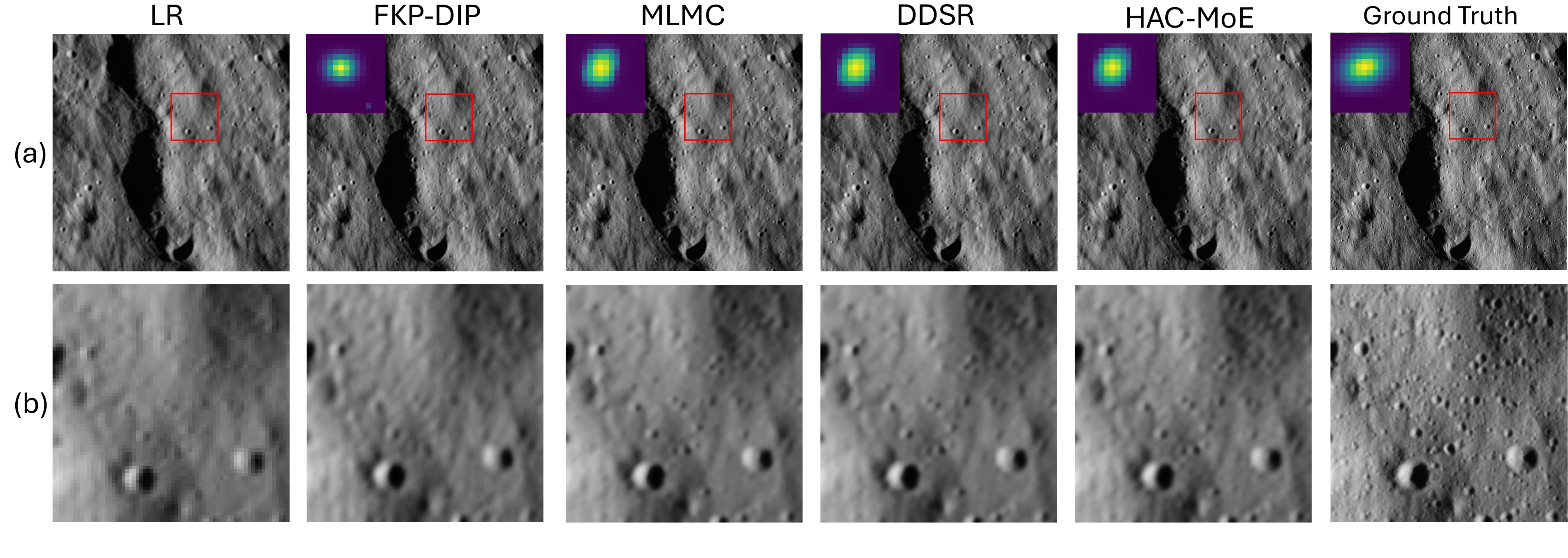}}
\label{fig_first_case}
\caption{Perceptual  comparisons in 3$\times$. The (a) include the ground truth, LR and reconstructed images, the estimated kernels are shown on top left. The (b) denote the zoomed-in for the select area from (a).}
\label{fig3x}
\end{figure*}

\begin{figure*}[!t]
\centering
{\includegraphics[width=7in]{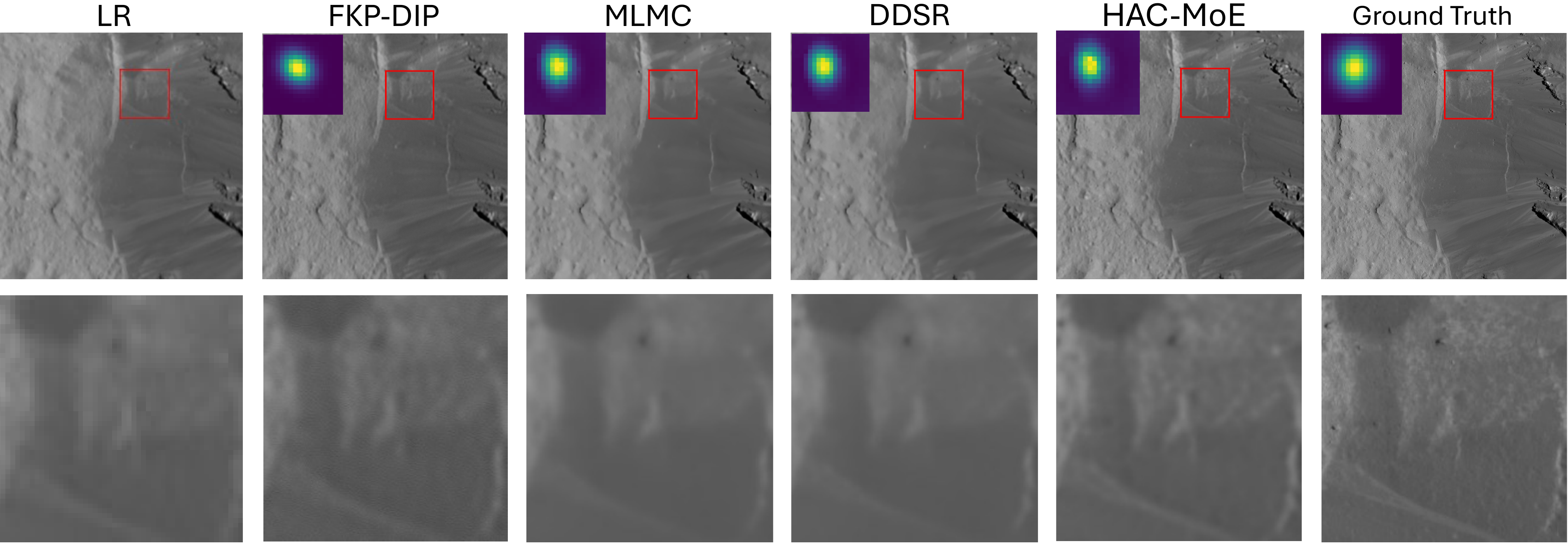}}
\label{fig_first_case}
\caption{Perceptual comparisons in 4$\times$.The (a) include the ground truth, LR and reconstructed images, the estimated kernels are shown on top left. The (b) denote the zoomed-in for the select area from (a). }
\label{fig4x}
\end{figure*}

\begin{figure*}[!t]
\centering
{\includegraphics[width=5.7in]{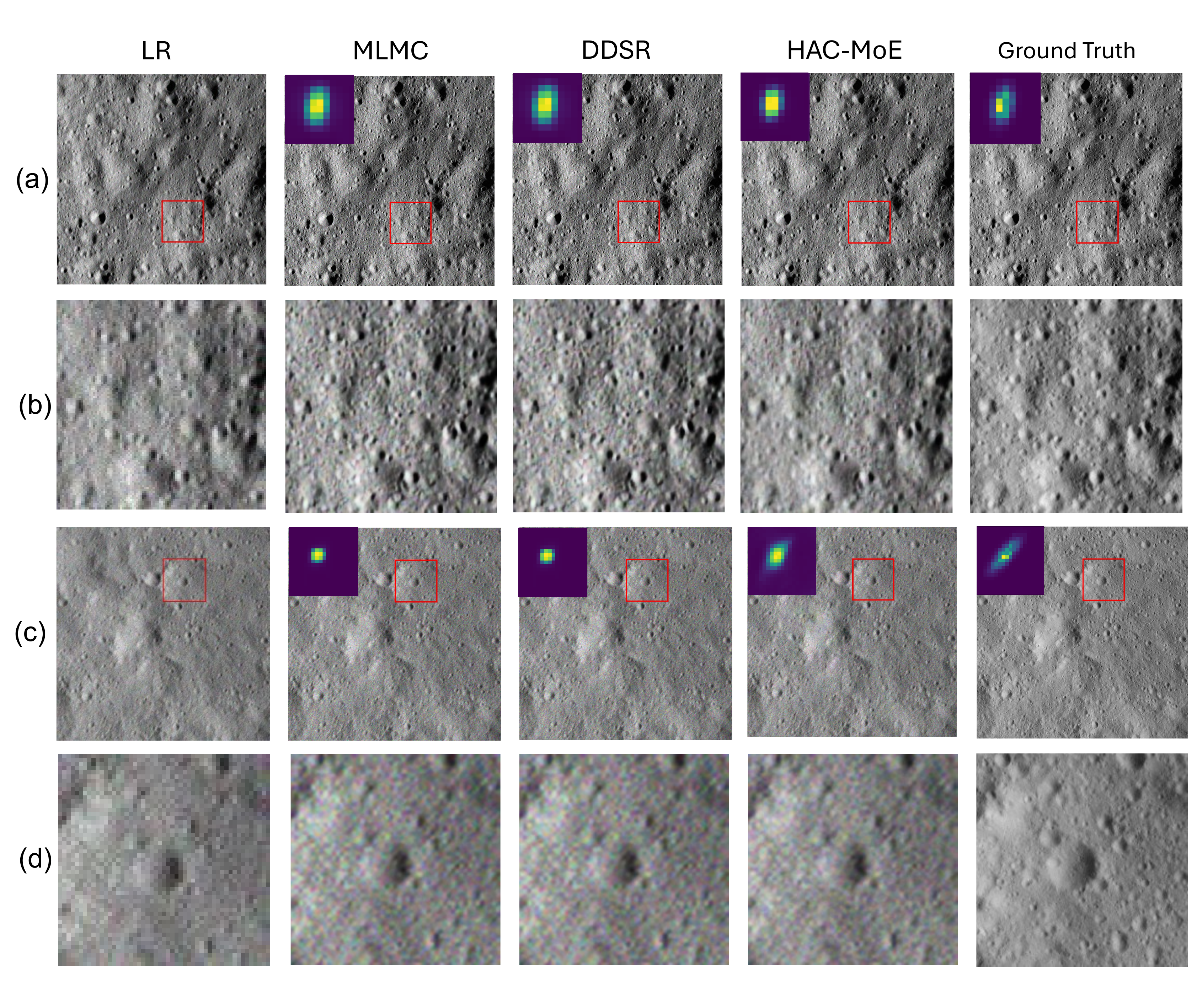}}
\label{fig_first_case}
\caption{Perceptual comparisons in 2$\times$ and 4$\times$ . The (a) (b) denote 2$\times$ SR and (c) (d) denote 4$\times$ SR. The (a) (c) include the ground truth, LR and reconstructed images, the estimated kernels are shown on top left. The (b) (d) denote the zoomed-in for the select area from (a) (c).}
\label{fig4x}
\end{figure*}

\section{Experiments}

\renewcommand{\arraystretch}{1.5}
\begin{table*}
\centering
\caption{Average PSNR/SSIM/LPIPS Results on Ceres-50 dataset under gaussian noise.  Best performance is shown in \textbf{bold}.}
\label{gaussian}
\begin{tabular}{p{2.5cm}<{\centering}|p{3.6cm}<{\centering}p{3.6cm}<{\centering}p{3.6cm}<{\centering}|p{2.5cm}<{\centering}}  
  \Xhline{1.2pt}
  \textbf{Method} & \textbf{$\times$2} & \textbf{$\times$3} & \textbf{$\times$4} & \textbf{Model Size} \\
  \Xhline{1.2pt}
  Bicubic  & 27.80/0.695/0.371 & 24.32/0.476/0.512& 23.76/0.445/0.591 & - \\
  FKP-DIP  & 29.87/0.731/0.390   & 28.97/0.675/0.471  & 27.90/0.650/0.541 & 2.5M \\
  DDSR  & 32.18/0.815/0.363   & 30.61/0.741/0.473   & 29.39/0.685/0.570 & 2.9M \\
  MLMC  & 32.12/0.815/0.366  & 30.60/0.740/0.473   & 29.40/0.686/0.578 & 2.9M \\
  HAC-MoE &   \textbf{32.28/0.816/0.304} & \textbf{30.95/0.759/0.407}  & \textbf{29.76/0.716/0.484}  & 3.1M \\
  \Xhline{1.2pt}
\end{tabular}
\end{table*}

\renewcommand{\arraystretch}{1.5}
\begin{table*}
\centering
\caption{Average PSNR/SSIM/LPIPS Results on Ceres-50 dataset under gaussian-POISSON noise.  Best performance is shown in \textbf{bold}.}
\label{gaussian_possion}
\begin{tabular}{p{2.5cm}<{\centering}|p{3.6cm}<{\centering}p{3.6cm}<{\centering}p{3.6cm}<{\centering}|p{2.5cm}<{\centering}}  
  \Xhline{1.2pt}
  \textbf{Method} & \textbf{$\times$2} & \textbf{$\times$3} & \textbf{$\times$4} & \textbf{Model Size} \\
  \Xhline{1.2pt}
  Bicubic  & 26.27/0.587/0.362  & 23.51/0.405/0.512 & 23.07/0.387/0.591 & - \\
  FKP-DIP  & 28.20/0.651/0.771   & 27.21/0.570/0.972  & 27.01/0.581/1.011 & 2.5M \\
  DDSR  & 28.50/0.660/0.748  &  27.52/0.592/0.960  &  27.02/0.579/1.008 & 2.9M \\
  MLMC  & 28.51/0.660/0.747  &  27.52/0.593/0.959  &  27.03/0.579/1.008 & 2.9M \\
  HAC-MoE & \textbf{29.41/0.695/0.559}  &  \textbf{28.03/0.624/0.781}  &  \textbf{27.14/0.583/0.974} & 3.1M \\
  \Xhline{1.2pt}
\end{tabular}
\end{table*}

\renewcommand{\arraystretch}{1.5}
\begin{table*}
\centering
\caption{Average PSNR/SSIM/LPIPS Results on Ceres-50 dataset for Ablation Analysis.  The notation -CK denotes removal of the Contrastive Kernel Sampling, -MoE denotes removal of the morphology-aware soft mixture-of-experts and -CL denotes removal of the History-augment Contrastive Learning component. Best performance is shown in \textbf{bold}.}
\label{Ablation_Study}
\begin{tabular}{p{4cm}<{\centering}|p{3cm}<{\centering}p{3cm}<{\centering}p{3cm}<{\centering}|p{2.5cm}<{\centering}}  
  \Xhline{1.2pt}
  \textbf{Method} & \textbf{$\times$2} & \textbf{$\times$3} & \textbf{$\times$4} & \textbf{Model Size} \\
  \Xhline{1.2pt}
  HAC-MoE            & 2\textbf{9.41/0.694/0.557 } &  \textbf{28.03/0.624/0.781}  &  \textbf{27.14/0.582/0.973 } &  3.1M\\
  HAC-MoE (-MoE)     & 29.29/0.684/0.559  &  27.96/0.617/0.786  &  27.09/0.581/0.975 & 2.9M\\
  HAC-MoE (-CK)(-CL)    & 29.24/0.673/0.561  &  27.99/0.620/0.796  &  27.10/0.580/0.976 &  3.1M\\
  HAC-MoE (-MoE)(-CK)(-CL)    & 29.19/0.665/0.571  &  27.93/0.613/0.798  &  27.06/0.579/0.980 &  2.9M\\
  \Xhline{1.2pt}
\end{tabular}
\end{table*}

\begin{figure*}[!t]
\centering
{\includegraphics[width=5.5in]{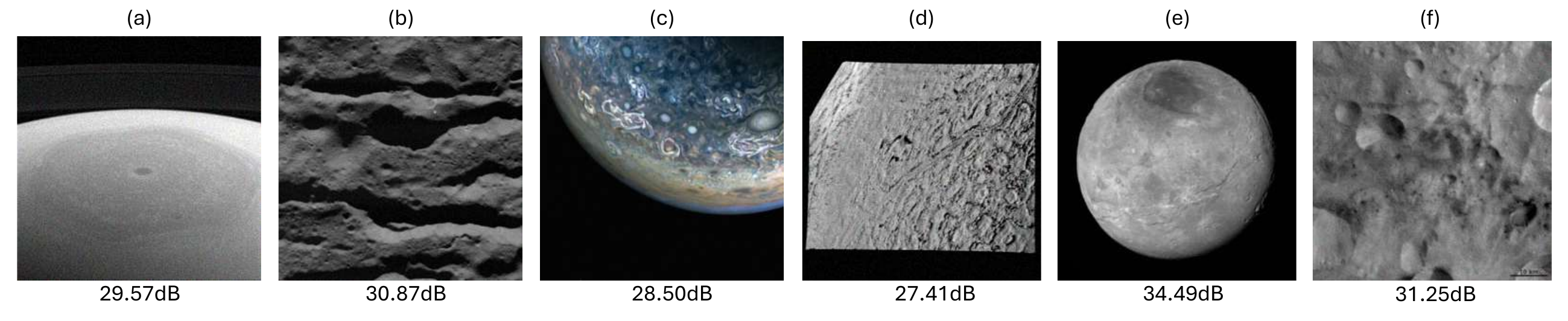}}
\caption{Generalization test of 4$\times$ BSR results for different morphological features on various planets. (a)  Saturn's northern hemisphere. (b) Deep Fractures in Occator Crater on Ceres. (c) Jupiter's surface. (d)  Northern Hemisphere of Neptune. (e) Charon's Complexity. (f)Dark Material Associated with and between Craters on Vesta.}
\label{example}
\end{figure*}

\subsection{Dataset and Data Preparation}

Unlike generic BSR methods that are primarily evaluated on natural image benchmarks, the proposed HAC-MoE is designed and validated specifically for planetary remote sensing scenarios. To support this objective, we construct the Ceres-50 dataset, which emphasizes diverse planetary surface morphologies and physically motivated degradation processes. The Ceres-50 dataset consists of 50 grayscale planetary images collected from NASA’s Small Bodies Node\footnote{\url{https://pdssbn.astro.umd.edu/}}. The images depict the surface of Ceres, a dwarf planet in the asteroid belt known for its rich geological diversity, including impact craters, fractures, bright deposits, and smooth regions \cite{Ceres_a_wet_planet,Dawn_arrives_at_Ceres}. These characteristics make Ceres a representative testbed for evaluating planetary image restoration methods. All ground-truth images have a spatial resolution of $1024 \times 1024$ pixels.

To generate low-resolution observations, we adopt a physically motivated degradation pipeline that reflects the imaging characteristics of deep-space planetary sensors. Specifically, each high-resolution image is first blurred using an anisotropic Gaussian kernel, followed by spatial downsampling and the addition of realistic sensor noise. The anisotropic Gaussian kernel has a size of $(4s+3) \times (4s+3)$, where $s$ denotes the downsampling factor. The standard deviations along the two principal axes are independently sampled from a uniform range $[0.175s,\, 2.5s]$, and the kernel is randomly rotated within $[0,\, \pi]$ to model directional blur caused by varying viewing geometries. To further account for residual optical aberrations and temporal instabilities, element-wise kernel perturbations sampled from $[-0.4,\, 0.4]$ are applied prior to kernel normalization. This process produces a diverse set of anisotropic point spread functions that approximate the spatial blurring behavior of deep-space framing cameras under different operational conditions \cite{The_Dawn_framing_camera}.

After blurring, the images are downsampled by a factor of $s$ to obtain low-resolution inputs. Sensor noise is then added using two complementary noise models to cover typical planetary imaging regimes. In the Gaussian noise setting, zero-mean Gaussian noise is added with a standard deviation randomly sampled from $(0,\, 0.039]$, corresponding to an intensity variation of approximately $\pm 10/255$. This setting approximates high signal-to-noise ratio (SNR) conditions dominated by signal-independent readout noise. In addition, we consider a Poisson--Gaussian noise model to simulate photon-limited imaging conditions. Given a normalized image $\mathbf{x} \in [0,1]$, the noisy observation $\tilde{\mathbf{x}}$ is generated as
\begin{equation}
\tilde{\mathbf{x}} = \alpha \, \mathrm{Poisson}\!\left(\frac{\mathbf{x}}{\alpha}\right) + \mathcal{N}\!\left(0,\, \sigma^2\right),
\end{equation}
where $\alpha$ denotes the effective sensor gain and $\sigma$ controls the standard deviation of the Gaussian noise. The $\mathcal{N}$ also denotes the gaussian noise. In our experiments, we set $\alpha = 3 \times 10^{-3}$ and $\sigma = 0.2$, following the imaging characteristics reported for the Dawn framing camera \cite{The_Dawn_framing_camera}. Together, the Gaussian and Poisson--Gaussian noise regimes cover a realistic range of operating conditions for planetary remote sensing instruments, from readout-noise-limited observations to photon-limited scenarios.

\subsection{Experiment details}

Experiments are conducted on an NVIDIA A100‑SXM4‑80GB GPU and an Intel Xeon Platinum 8168 CPU (16 cores), running Ubuntu 22.04.4 LTS, Python 3.12, and PyTorch 2.5.0 with CUDA 12.6. HAC-MoE is optimized for $N$ 300 iterations. The learning rates for the MA-MoE image estimator and the kernel generator MLP are set to 0.005 and 0.5, respectively. During the Contrastive Image Learning stage, parameters are updated every 10 iterations. In each iteration, HAC-MoE processes a single image and performs unsupervised learning on the Ceres‑50 images sequentially and each time conduct BSR on single image. For quantitative results, LPIPS uses AlexNet as the encoder. The inner loop of $P$ is set to 5 and the outer loop of $N$ is set to 120 for the gaussian noise. The number of experts is 4.

\noindent \textbf{Quantitative results.}   

We compare HAC-MoE with representative unsupervised BSR baselines, including DDSR~\cite{DDSR}, MLMC~\cite{MLMC}, and FKP-DIP~\cite{liang21FKP_DIP}. The results in Table~\ref{gaussian} and Table~\ref{gaussian_possion} show that HAC-MoE achieves competitive performance across $\times2$, $\times3$, and $\times4$ upscaling factors in terms of PSNR/SSIM/LPIPS. These gains are consistent with the design of HAC-MoE, which integrates (i) contrastive kernel sampling to reduce the distribution bias and suppress outlier kernels during kernel learning, (ii) history-augmented contrastive regularization to stabilize the unsupervised optimization in the image learning stage, and (iii) the morphology-aware soft mixture-of-experts (MA-MoE) image estimator to better accommodate heterogeneous planetary surface structures. Together, these components improve reconstruction fidelity under the Ceres-50 degradation setting.

\noindent \textbf{Qualitative results.}  

Figures~\ref{fig2x}, \ref{fig3x}, and \ref{fig4x} present visual comparisons of different methods at 2$\times$, 3$\times$, and 4$\times$ upscaling factors, respectively. The proposed HAC-MoE method demonstrates superior performance in both kernel reconstruction and image reconstruction across all scaling factors. This performance advantage is particularly evident at 4$\times$ upscaling, where HAC-MoE maintains sharper edge preservation and produces more accurate kernel estimations compared to competing methods.

\subsection{Ablation Study} 

To validate the performance of key components in HAC-MoE, ablation experiments were conducted. The experimental results are presented in Table~\ref{Ablation_Study}. The results demonstrate that removing the Contrastive Kernel Sampling significantly reduces PSNR and SSIM performance, particularly at lower BSR scales. Conversely, removing the Contrastive Learning (CL), morphology-aware soft mixture-of-experts (MoE) and Contrastive Kernel Sampling (CK) components adversely affects reconstruction quality at higher BSR scales, with the most pronounced impact observed at $2\times$ scale. These experimental findings confirm that both the Contrastive Kernel Generator and Contrastive Learning components play critical roles in the overall performance of HAC-MoE.

\section{Conclusion}

This paper presents HAC-MoE, an unsupervised framework for BSR in planetary remote sensing that operates without ground-truth high-resolution images or predefined degradation kernel priors. HAC-MoE integrates a kernel generator trained with contrastive kernel sampling and a MoE based image estimator optimized under observation consistency. Through alternating optimization, the framework stabilizes both kernel estimation and image reconstruction in a fully unsupervised setting.

Experimental results on the Ceres-50 dataset demonstrate that HAC-MoE achieves competitive performance among existing unsupervised BSR methods under anisotropic Gaussian blur with kernel perturbations, physically motivated Gaussian and Gaussian-Poisson sensor noise, and spatial downsampling. These results validate the effectiveness of the proposed kernel learning strategy as well as the history-augmented contrastive image learning mechanism for planetary images characterized by unknown and heterogeneous degradations.

Despite HAC-MoE achieve impressive performance, the degradation models considered in this work remain limited in scope. Real planetary remote sensing data may exhibit additional artifacts such as striping noise, sensor-induced distortions, missing-data masks, and other non-ideal acquisition effects that are not explicitly modeled in the current framework. Future work will focus on extending HAC-MoE to accommodate broader and more complex degradation families, as well as incorporating mechanisms to handle occlusions and incomplete observations. In addition, further exploration of contrastive learning strategies and adaptive kernel modeling may improve robustness and reconstruction quality under more challenging planetary imaging conditions.





\bibliographystyle{IEEEtran}
\bibliography{reference/blind_super_resolution, reference/contrastive,reference/zhj_work, reference/space_science, reference/super_resolution, reference/tscvt, reference/moe}
%

\vspace{11pt}

\begin{IEEEbiography}[{\includegraphics[width=1in,height=1.25in,clip,keepaspectratio]{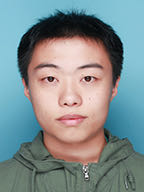}}]{Hui-Jia Zhao}
received the B.S. degree  from Henan University, Kaifeng, China, in 2022. He is currently pursuing the Ph.D. degree in computer technology and application with Macau University of Science and Technology, Macau, China.

His research interests include deep learning and super-resolution.
\end{IEEEbiography}

\begin{IEEEbiography}[{\includegraphics[width=1in,height=1.25in,clip,keepaspectratio]{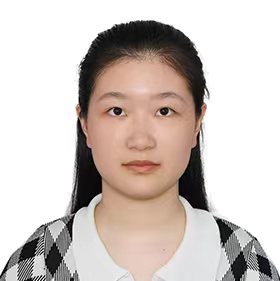}}]{Jie Lu}
received the B.S. degree in computer science in 2023 and is currently pursuing the M.S. degree in Applied Mathematics and Data Science from Macau University of Science and Technology, Macau, China.

Her research interests include deep learning and contrastive learning.
\end{IEEEbiography}

\begin{IEEEbiography}[{\includegraphics[width=1in,height=1.25in,clip,keepaspectratio]{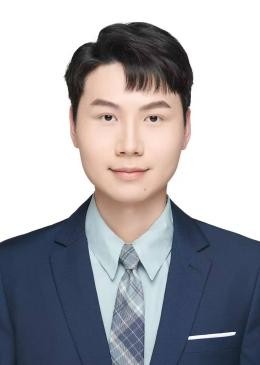}}]{Yunqing Jiang}
Yunqing Jiang is a Ph.D. candidate at the Macau University of Science and Technology. His research interests include open-vocabulary object detection, knowledge distillation based on large models, controllable generation, and robot policy. He has published a first-author paper in Knowledge-Based Systems and co-authored a paper in the Journal of Statistical Computation and Simulation. He is currently working as an intern at the Guangdong Institute of Intelligence Science and Technology.
\end{IEEEbiography}

\begin{IEEEbiography}[{\includegraphics[width=1in,height=1.25in,clip,keepaspectratio]{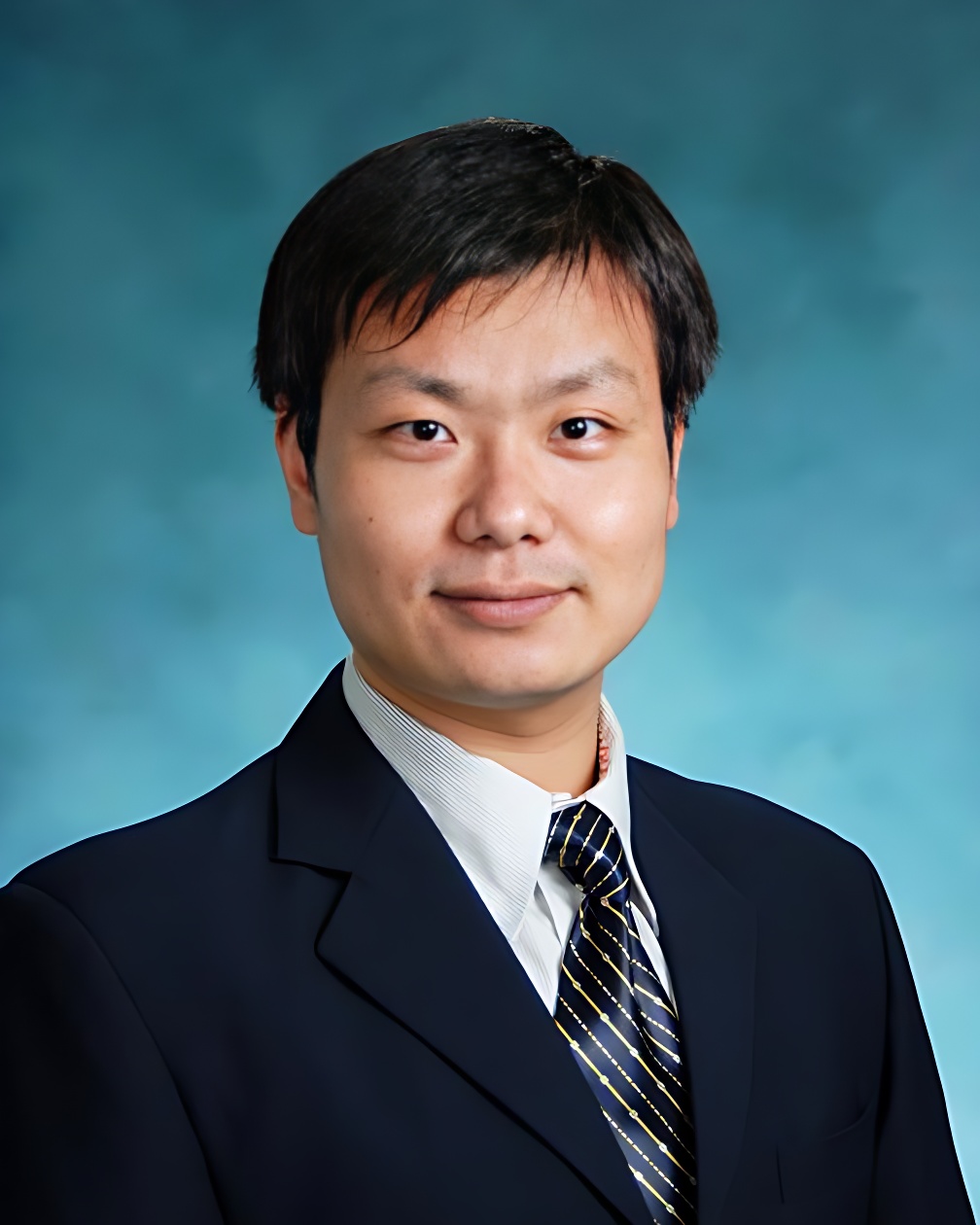}}]{Xiao-Ping Lu}
received the B.S. degree in information and computing science and M.S. degree in computational mathematics in 2004 and 2007, respectively, from Fudan University, Shanghai, China, and the Ph.D. degree in computer technology and application in Macau University of Science and Technology in 2013. He had a short-term visiting in Tampere University of Technology, Finland, and was a scholar visitor in UCLA, USA. 

He is currently a Professor with the School of Computer Science and Engineering, Faculty of Innovation Engineering in Macau University of Science and Technology, Macau, China, where he is also with the State Key Laboratory of Lunar and Planetary Sciences. His research interests include machine learning and its applications (deep learning, reinforcement learning); data science; high performance computing, and remote sensing data processing and analysis.
\end{IEEEbiography}

\begin{IEEEbiography}[{\includegraphics[width=1in,height=1.25in,clip,keepaspectratio]{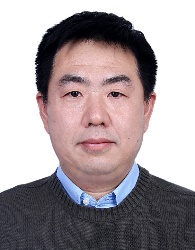}}]{Kaichang Di}
received the Ph.D. degree in photogrammetry and remote sensing from Wuhan Technical University of Surveying and Mapping, Wuhan, China, in 1999. 
He is currently a Professor with Aerospace Information Research Institute, Chinese Academy of Sciences, Beijing, China. And he is the founding director of the Planetary Mapping and Remote Sensing Lab at the institute. His research interests include planetary photogrammetry and remote sensing, visual localization and navigation, and planetary science.

\end{IEEEbiography}


\vspace{11pt}

\vfill

\end{document}